\documentclass[letterpaper, 10 pt, conference]{ieeeconf}  

\IEEEoverridecommandlockouts                              
\usepackage{graphicx}
\usepackage{amsmath, amssymb}
\usepackage{algorithm}
\usepackage{algpseudocode}
\usepackage{xcolor}
\usepackage{booktabs} 
\usepackage{multirow}
\usepackage{caption}
\DeclareCaptionLabelFormat{alglabel}{Alg.~#2}
\DeclareCaptionLabelFormat{tablabel}{Tab.~#2}
\usepackage{hyperref}
\usepackage{dblfloatfix}
\usepackage{afterpage}
\usepackage{tikz} 
\usetikzlibrary{calc, decorations.pathreplacing} 
\newcommand{\tikzmark}[1]{\tikz[overlay, remember picture] \node (#1) {};}

\usepackage{hyperref}
\hypersetup{
  colorlinks = true,  
  urlcolor   = deeppink,  
  linkcolor  = blue,  
  citecolor  = red,  
  pdfview    = XYZ,  
}

\hypersetup{colorlinks,allcolors=[rgb]{0,0.07843,0.45098}, urlcolor=blue}

\definecolor{customred}{HTML}{A31F34}

\usepackage{tabularx}
\newcommand{\GD}{\textit{GD}}
\newcommand{\BI}{\textit{BI}}
\newcommand{\RS}{\textit{RS}}
\newcommand{\OP}{\textit{OP}}
\newcommand{\SGS}{\textit{SS}}
\newcommand{\PR}{\textit{PR}}

\title{\LARGE \bf
Inference-Time Policy Steering through Human Interactions
}

\author{
    Yanwei Wang$^{\dag}\textsuperscript{*}$, Lirui Wang$^{\dag}$, Yilun Du$^{\dag}$, Balakumar Sundaralingam$^{\ddag}$, 
    Xuning Yang$^{\ddag}$, Yu-Wei Chao$^{\ddag}$, \\ Claudia Pérez-D’Arpino$^{\ddag}$, Dieter Fox$^{\ddag}$, Julie Shah$^{\dag}$%
     }
\begin{document}
\maketitle

\thispagestyle{empty}
\pagestyle{empty}


\begin{abstract}
Generative policies trained with human demonstrations can autonomously accomplish multimodal, long-horizon tasks. However, during inference, humans are often removed from the policy execution loop, limiting the ability to guide a pre-trained policy towards a specific sub-goal or trajectory shape among multiple predictions. Naive human intervention may inadvertently exacerbate distribution shift, leading to constraint violations or execution failures. To better align policy output with human intent without inducing out-of-distribution errors, we propose an Inference-Time Policy Steering (ITPS) framework that leverages human interactions to bias the generative sampling process, rather than fine-tuning the policy on interaction data. We evaluate ITPS across three simulated and real-world benchmarks, testing three forms of human interaction and associated alignment distance metrics. Among six sampling strategies, our proposed stochastic sampling with diffusion policy achieves the best trade-off between alignment and distribution shift. Videos are available at \href{https://yanweiw.github.io/itps/}{https://yanweiw.github.io/itps/}.
\vspace{-1pt}
\begingroup
\renewcommand\thefootnote{}
\footnote{$^{\dag}$MIT CSAIL, $^{\ddag}$NVIDIA. \textsuperscript{*}Partly completed during NVIDIA internship.}
\setcounter{footnote}{0} 

\endgroup

\end{abstract}

\section{Introduction}

Behavior cloning \cite{osa2018algorithmic} has fueled a recent wave of generalist policies \cite{o2023open, team2024octo, kim2024openvla} capable of solving multiple tasks using a single deep generative model \cite{urain2024deep}. As these models acquire an increasing number of dexterous skills \cite{chi2023diffusion, zhao2023learning, fu2024mobile} from multimodal\footnote{In this work, multimodal refers to the data distribution, not interaction or sensor modalities.} human demonstrations, the natural next question arises: how can these skills be tailored to follow specific user objectives? Currently, there are few mechanisms to directly intervene and correct the behavior of these out-of-the-box policies at inference time, particularly when their actions misalign with user intent---often due to task underspecification or distribution shift during deployment.

\begin{figure}
    \includegraphics[width=\linewidth]{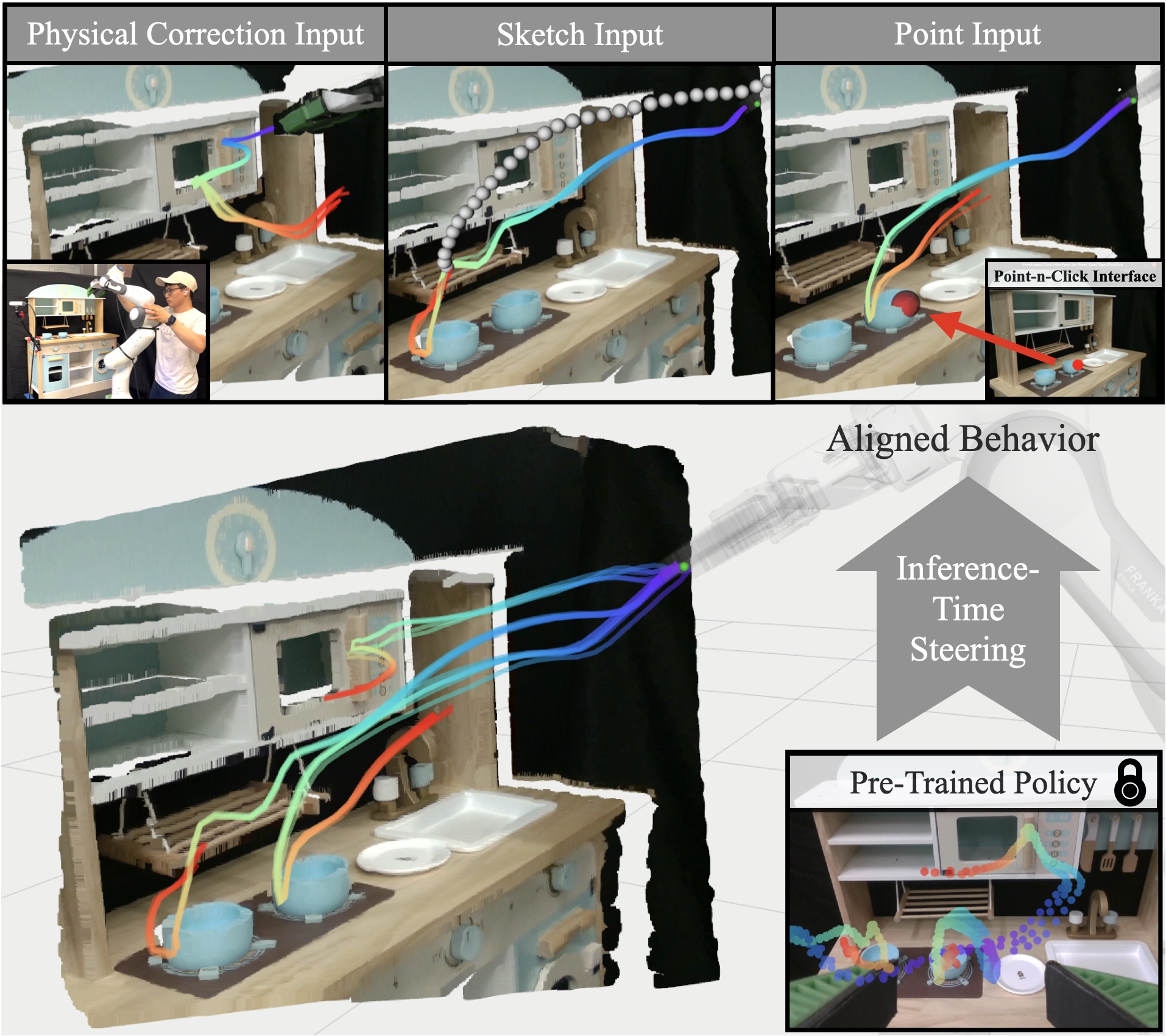}
    \captionof{figure}{\small \textbf{Inference-Time Policy Steering (ITPS).} We present a novel framework to unify various forms of human interactions to steer a frozen generative policy. User interactions ``prompt'' pre-trained policies to synthesize aligned behaviors at inference time. }
    \label{fig:framework}
    \vspace{-12pt}
\end{figure}

One strategy for adapting policies designed for autonomous behavior generation to real-time human-robot interaction is to fine-tune them on interaction data, such as language corrections \cite{shi2024yell}. However, this approach requires additional data collection and training, and language may not always be the best modality for capturing low-level, continuous intent \cite{gu2023rt}. In this work, we explore whether a frozen pre-trained policy can be \textit{steered to generate behaviors aligned with user intent}---specified directly in the task space through point goals \cite{kemp2008point}, trajectory sketches \cite{gu2023rt}, and physical corrections \cite{wang2024grounding} (Figure \ref{fig:framework})---without fine-tuning.

While inference-time interventions in the task space offer a direct way to guide behavior, they can inadvertently exacerbate distribution shift---a well-known issue in behavior cloning that often leads to execution failures \cite{ross2011reduction}.
Prior works addressing this issue \cite{losey2018review, losey2022physical, billard2022learning, wang2022temporal} largely focus on single-task settings, limiting their applicability to multi-task policies. To overcome this limitation, we leverage multimodal generative models to produce trajectories that respect likelihood constraints \cite{janner2022planning, ajay2022conditional, ye2024efficient}, ensuring the policy generates valid actions even after steering. Specifically, we frame policy steering as conditional sampling from the likelihood distribution of a learned generative policy. The likelihood constraints learned from successful demonstrations allow us to consistently synthesize valid trajectories, while conditional sampling ensures that these trajectories align with user objectives. By composing pre-trained policies with inference-time objectives, we can flexibly adapt generalist policies to each new downstream interaction modality, {\it without needing to modify the pre-trained policy in any way.} 

\begin{figure*}[t!] 
  \begin{minipage}[t]{0.64\textwidth} 
    \raggedleft
    \includegraphics[width=\linewidth]{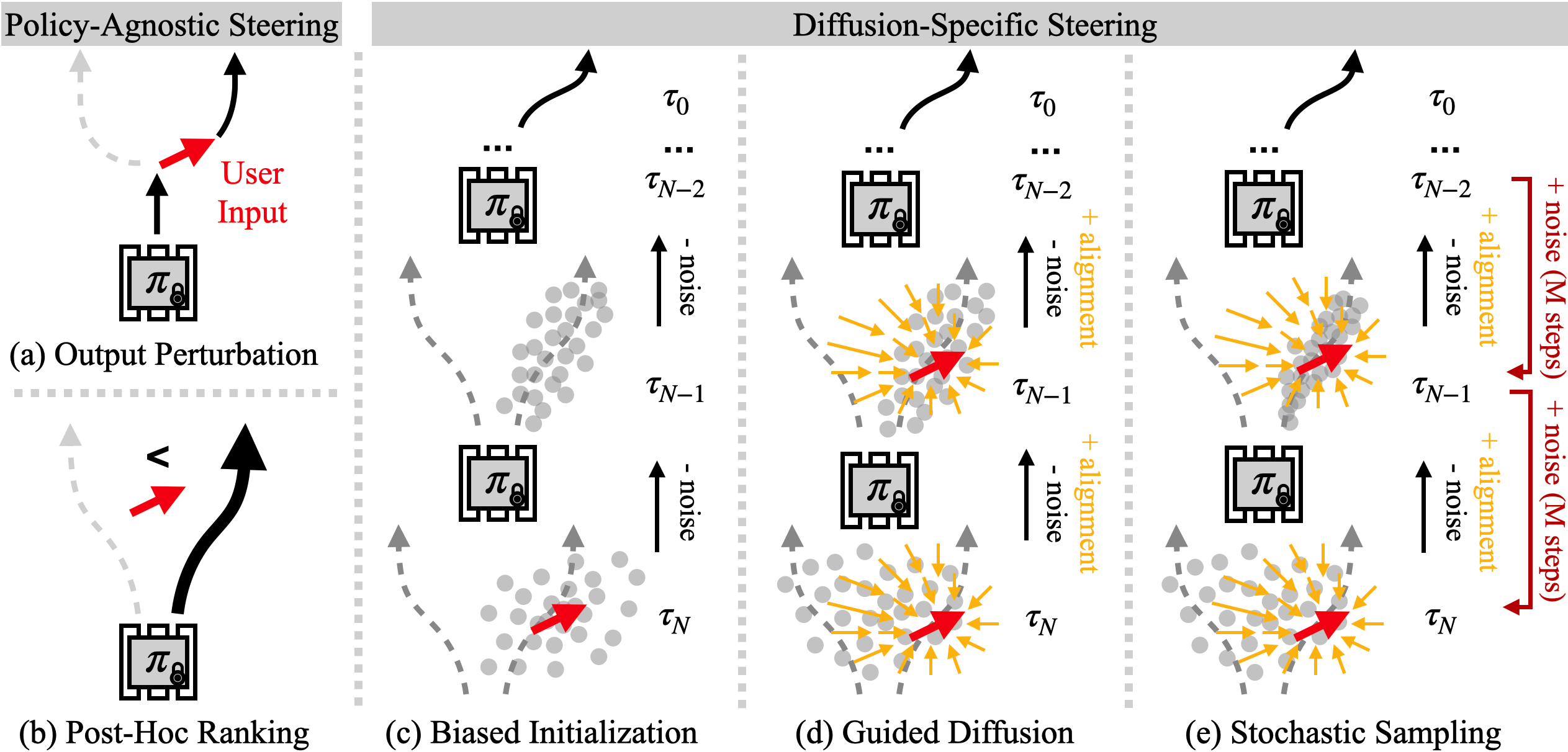} 
    \caption{\small \textbf{Policy Steering Methods.} Given user input, methods (a-c) incorporate the alignment objective either before or after inference via (a) perturbation, (b) ranking, or (c) initialization, whereas methods (d,e) integrate the objective directly during inference.}
    \label{fig:method}
  \end{minipage} \hfill
  \begin{minipage}[t]{0.34\textwidth} 
  \vspace*{-158pt}
    \small
    \captionsetup{width=0.90\linewidth}
    \begin{tabular}{p{1\linewidth}} 
        \rule{\linewidth}{1pt}  
        \vspace{-0.3em}
        \noindent\textbf{\hspace{1em} Algorithm 1: Stochastic Sampling} \\[-0.3em]
        \rule{\linewidth}{0.5pt}  
        \noindent\textbf{\hspace{1em} Input:} diffusion policy $\pi_\theta$, user interaction $\mathbf{z}$, alignment objective $\xi(\cdot)$ \\
        1: Initialize plan $\tau_N \sim \mathcal{N}(0, I)$ \\
        2: \textbf{for} $i = N, \dots, 1:$ \textcolor{gray}{// denoising steps} \\
        3: \hspace{1em} \textcolor{customred}{\textbf{for} $j = 1, \dots, M:$} \textcolor{customred}{// sampling steps} \\
        4: \hspace{2em} $\epsilon \gets \pi_\theta(\tau_i)$ \\
            \textcolor{gray}{\hspace{4em} // denoising gradient} \\
        5: \hspace{2em} $\delta \gets \nabla \xi(\tau_i, \mathbf{z})$ \\
            \textcolor{gray}{\hspace{4em} // alignment gradient} \\
        6: \hspace{2em} $\textcolor{customred}{\textbf{if } j < M}$: \\
        7: \hspace{3em} $\textcolor{customred}{\tau_{i} \gets \texttt{reverse}(\tau_i, \epsilon + \beta_i \delta, i)}$ \\
        8: \hspace{2em} \textcolor{customred}{\textbf{else}}: \\
        9: \hspace{3em} $\tau_{i-1} \gets \texttt{reverse}(\tau_i, \epsilon + \beta_i \delta, i-1)$ \\
        \rule{\linewidth}{0.5pt}  
    \end{tabular}
    \vspace*{-3pt}
    \captionof{algorithm}{\small \textbf{Stochastic Sampling.}  \textcolor{customred}{A four-line change} from a guided diffusion algorithm.}
    \label{alg:psuedocode}
  \end{minipage}
  \vspace{-15pt}
\end{figure*}

To evaluate the effectiveness of inference-time steering, we formulate discrete and continuous alignment metrics to capture human preferences in discrete task execution and continuous motion shaping. We study a suite of six methods for converting interaction inputs into conditional sampling on generative models. We identify an alignment-constraint satisfaction trade-off: as these methods improve alignment, they tend to produce more constraint violations and task failures. To address this, we propose an MCMC procedure \cite{du2023reduce} for diffusion policy \cite{chi2023diffusion} that alleviates distribution shift during interaction-guided sampling, achieving the best alignment-constraint satisfaction trade-off across various combinations of generative policies and sampling strategies.

\textbf{Contributions} \textbf{(1)} We propose a novel inference-time framework (ITPS) that incorporates real-time user interactions to steer frozen imitation policies. \textbf{(2)} We introduce a set of alignment objectives, along with sampling methods for optimizing these objectives, and illustrate the alignment-constraint satisfaction trade-off. \textbf{(3)} We design a new inference algorithm for diffusion policy—stochastic sampling—which improves sample alignment with user intent while maintaining constraints within the data manifold.
\vspace{-10pt}

\section{Policy Steering}
\subsection{Steering Towards User Intent}
In this work, we explore how to produce trajectories $\tau$ from frozen generative models that align with user intent specified either as discrete tasks (e.g. picking left or right bowl as shown in Figure 1) or continuous motions. For discrete preferences, we aim to maximize \textbf{T}ask \textbf{A}lignment (\texttt{TA}) as the percentage of predicted skills that execute intended tasks. For continuous preferences, we aim to maximize \textbf{M}otion \textbf{A}lignment (\texttt{MA}) as the negative $\mathcal{L}_2$ distance between generated trajectories and target trajectories. In addition to explicitly specified user objectives, we measure the percentage of generated plans that satisfy physical constraints---implicit user intents such as avoiding collisions or completing tasks---referred to as the \textbf{C}onstraint \textbf{S}atisfaction rate (\texttt{CS}). We define {\it steering towards user intent} as increasing \texttt{TA} or \texttt{MA} while maximizing \texttt{CS}. Specifically, maximizing \texttt{CS} is achieved through sampling in distribution of a pre-trained policy, while increasing \texttt{TA} or \texttt{MA} is achieved through minimizing an objective function $\xi(\tau, \mathbf{z})$, where user informs his intent through interactions $\mathbf{z}$ to score the space of trajectories $\tau$. 
We consider the following three interaction types and objective functions.

\textbf{Point Input.}  The first objective function $\xi$ has a user specify a point coordinate on an image we wish to have a robot trajectory reach. Given a generated trajectory $\tau = (\mathbf{s}_1, \mathbf{s}_2, \dots, \mathbf{s}_T) \in \mathbb{R}^3$, we map the specified pixel using the depth information in an RGB-D scene camera to a corresponding 3D state $\mathbf{z^{\text{point}}} \in \mathbb{R}^3$. The alignment to user intent is then defined as minimizing the objective function:
\begin{equation}
\xi(\tau, \mathbf{z}^{\text{point}}) = \sum_{t=1}^{T} \frac{1}{T} \|\mathbf{s}_t - \mathbf{z}^{\text{point}} \|_2, 
\end{equation}
which captures the average $\mathcal{L}_2$-distance between all states in the generated trajectory and the target 3D state $\mathbf{z}$\footnote{While $\min\limits_{s_1 \dots s_T} \|\mathbf{s}_t - \mathbf{z}\|_2$ is more accurate, gradients are not smooth.}. 
This objective function allows users to flexibly point goals in a scene, by specifying which objects to manipulate in a real-world kitchen environment (Figure \ref{fig:framework}).

\textbf{Sketch Input.} The next objective function $\xi$ we consider allows a user to specify a more continuous intent, by generating a partial trajectory sketch $\mathbf{z^\text{sketch}} \in \mathbb{R}^{T \times 3}$ that we wish to have the robot follow. Given this sketch, we define $\xi$ as:
\begin{equation}
\label{eq:sketch_input}
\xi(\tau, \mathbf{z}^\text{sketch}) = \sum_{t=1}^{T} \|\mathbf{s}_t - \mathbf{z}^\text{sketch}_t\|_2.
\end{equation}
When the sketch $\mathbf{z}$ has a different length than generated trajectories $\tau$, we uniformly resampled $\mathbf{z}^\text{sketch}$ to match the temporal dimension of generated samples\footnote{$\mathcal{L}_2$ used over DTW \cite{muller2007dynamic} for smooth gradients and linear time complexity.}.
In comparison to the point input, this objective function allows users to specify shape preferences of a trajectory through a directional path in a robot's workspace (Figure \ref{fig:maze2d_qualitative}).

\textbf{Physical Correction Input.} Finally, we consider an objective $\xi$ which allows a user to specify intent through physical corrections $\mathbf{z}^{\text{nudge}}$ on the robot. Minimizing the objective
\begin{equation}
\xi(\tau, \mathbf{z}^{\text{nudge}}) = 
\begin{cases}
     0, & \mathbf{s}_t = \mathbf{z}^{\text{nudge}}_t \text{ for } t \le k \\
     \infty, & \text{otherwise}
\end{cases}
\end{equation}
corresponds to overwriting the beginning portion (e.g. first $k$ steps) of a trajectory $\tau$ with a user-specified $\mathbf{z^{\text{nudge}}}$:
\begin{equation}
\tau = [\mathbf{z}^{\text{nudge}}_1, \dots, \mathbf{z}^{\text{nudge}}_k,  {\textbf{s}}_{k+1}, \dots,  {\textbf{s}}_T].
\end{equation}
Compared to previous interaction types, physical corrections intervene directly in the robot's motion execution (Figure \ref{fig:framework}).

\begin{figure*}[t!] 
  \begin{minipage}[t]{0.58\linewidth}
    \includegraphics[width=\linewidth]{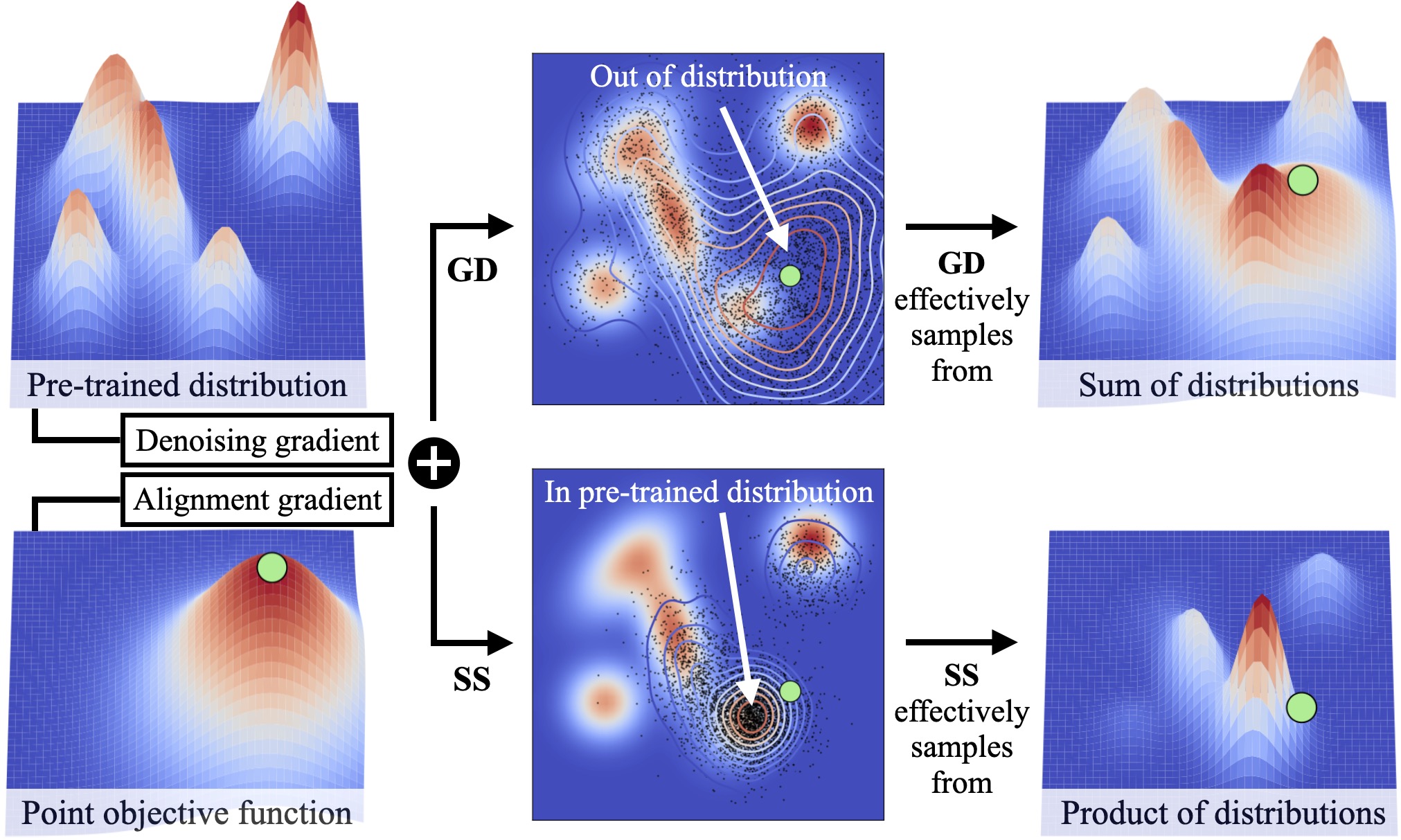}
    \captionof{figure}{\small \textbf{Guided Diffusion vs. Stochastic Sampling.} In a toy example aiming to sample likely data points from a pre-trained distribution while aligning with a target point, \GD{} samples approximate the sum of two distributions, whereas \SGS{} samples approximate their product, as illustrated by contour lines from kernel density estimation \cite{Waskom2021}. Consequently, when the point input does not align with any distribution mode, \GD{} introduces distribution shift, while \SGS{} identifies the closest in-distribution mode.}
    \label{fig:gd_vs_ss}
  \end{minipage}
  \hfill
  \begin{minipage}[t]{0.40\linewidth}
    \raggedright
    \vspace{-162pt}
        \includegraphics[width=0.99\linewidth]{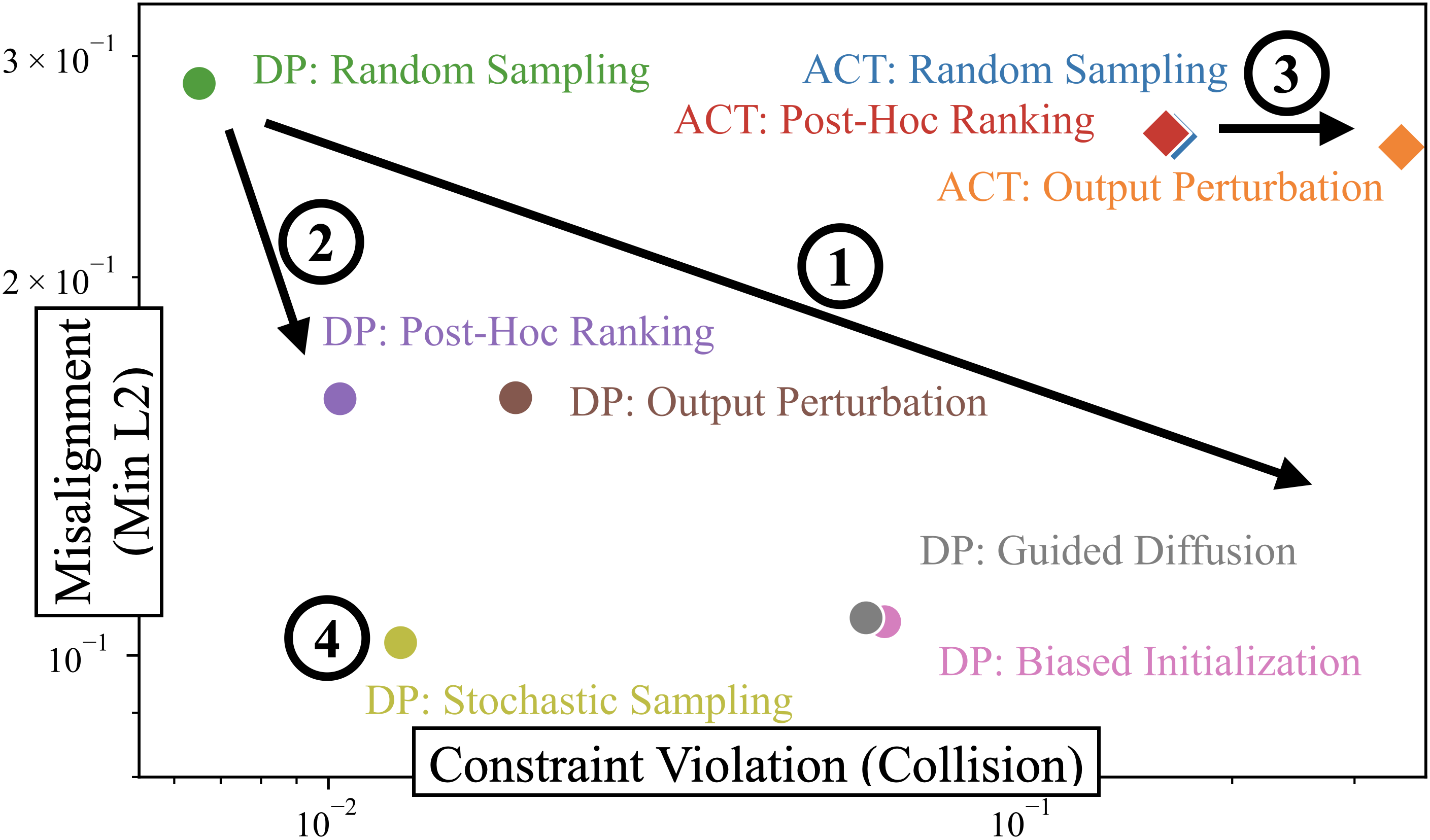}
        \captionsetup{width=0.96\linewidth}
        \caption{\small \textbf{Alignment vs. Collision in Maze2D.} 
        We compare various sampling methods with ACT and DP steered using sketch input. (1) Steering frozen policies improves alignment at the cost of constraint satisfaction and increased collisions. Moreover, (2) Multimodal policies (DP) steered with \PR{} enhance alignment without significant distribution shift, while (3) unimodal policies (ACT) are harder to steer effectively, particularly if they lack robustness (see Figure \ref{fig:maze_csr}). (4) Finally, DP steered with \SGS{} achieves the best alignment-constraint satisfaction trade-off.
    \label{fig:maze_multi_trend}}
  \end{minipage}
    \vspace{-20pt} 
\end{figure*}

\subsection{Inference-Time Interaction-Conditioned Sampling}

Given an inference-time alignment objective $\xi(\tau, \mathbf{z})$ on trajectories $\tau$, we explore six methods for biasing trajectory generation to minimize this objective. The first three methods are applicable across generative models parameterized by $\theta$, while the latter three specifically leverage the implicit optimization procedure within the diffusion process. Figure~\ref{fig:method} illustrates these optimization procedures.

\textbf{Random Sampling (RS).} In the \textit{Random Sampling} baseline, we sample a trajectory $\tau \sim \pi_\theta$ directly from the pre-trained model without any modification. This approach does not explicitly optimize any objective function $\xi$, but serves as a baseline for trajectory generation.

\textbf{Output Perturbation (\OP).} In \textit{Output Perturbation}, we first sample a trajectory $\tau$ from $\pi_\theta$ and apply a post-hoc perturbation to minimize the objective $\xi(\tau, \mathbf{z}^{\text{nudge}})$. We then resample from $\mathbf{z}_k^{\text{nudge}}$ to complete the remainder of trajectory $\tau$. If a user cannot provide direct physical correction, the first $k$ states of a sketch input can be used as $\mathbf{z}^{\text{nudge}}$. Although this sampling strategy maximizes alignment up to step $k$, it does not guarantee that synthesized trajectories from the perturbed state $\mathbf{z}_k^{\text{nudge}}$ will be constraint satisficing.

\textbf{Post-Hoc Ranking (PR).} In \textit{Post-Hoc Ranking}, we generate a batch of $N$ trajectories $\{\tau_j\}_{j=1}^N$ from $\pi_\theta$ and select $\tau^*$ that minimizes the objective $\xi(\tau, \mathbf{z}^{\text{point}})$ or $\xi(\tau, \mathbf{z}^{\text{sketch}})$.
This approach performs well when at least one generated sample closely aligns with the input $\mathbf{z}$, which may not hold if the robot is in a state without multimodal policy predictions.

\textbf{Biased Initialization (BI).}
In \textit{Biased Initialization}, inspired by \cite{yoneda2023noise}, we  modify the initialization of the reverse diffusion process. Instead of initializing with a noise trajectory $\tau_N$\footnote{Subscript denotes diffusion steps for $\tau_i$ and trajectory timesteps for $s_t$.} $\sim \mathcal{N}(0, I)$, we use a Gaussian-corrupted version of the user input $\mathbf{z^\text{point}}$ or $\mathbf{z^\text{sketch}}$ as $\tau_N$, bringing the process closer to the desired mode from the outset.
While this approach specifies user intent at initialization, the sampling process may still deviate from this input.

\textbf{Guided Diffusion (\GD).}
In \textit{Guided Diffusion}, we use the objective function $\xi(\tau, \mathbf{z})$ to guide the trajectory synthesis in the diffusion process~\cite{janner2022planning}. Specifically, at each diffusion timestep $i$, given $\mathbf{z^\text{point}}$ or $\mathbf{z^\text{sketch}}$, we compute the alignment gradient $\nabla_{\tau_i} \xi(\tau_i, \mathbf{z})$ to bias sampling:
\begin{equation}
\tau_{i-1} = \alpha_i(\tau_i - \gamma_i (\epsilon_\theta(\tau_i, i) + \beta_i \nabla_{\tau_i} \xi(\tau_i, \mathbf{z}))) + \sigma_i \eta,
\label{eqn:guided_diffusion}
\end{equation}
 where $\epsilon_\theta(\tau_i, i)$ is the denoising network, $\eta \sim \mathcal{N}(0, I)$ is Gaussian noise, $\beta_i$ is the guide ratio that controls the alignment gradient's influence, $\alpha_i$, $\gamma_i$, $\sigma_i$ are diffusion-specific hyperparameters. This alignment gradient steers the reverse process toward trajectories aligned with $\mathbf{z}$, potentially discovering new behavior modes in states where unconditional predictions would otherwise be unimodal and far from  $\mathbf{z}$. However, sampling with a weighted sum of denoising and alignment gradients in Equation~\ref{eqn:guided_diffusion} approximates sampling from the weighted sum of the policy distribution and the objective distribution rather than their product~\cite{du2023reduce}, which can result in out-of-distribution samples (Figure \ref{fig:gd_vs_ss}).

 \begin{figure*}[t] 
  \begin{minipage}[t]{0.60\textwidth} 
    \raggedleft
    \includegraphics[width=1\linewidth]{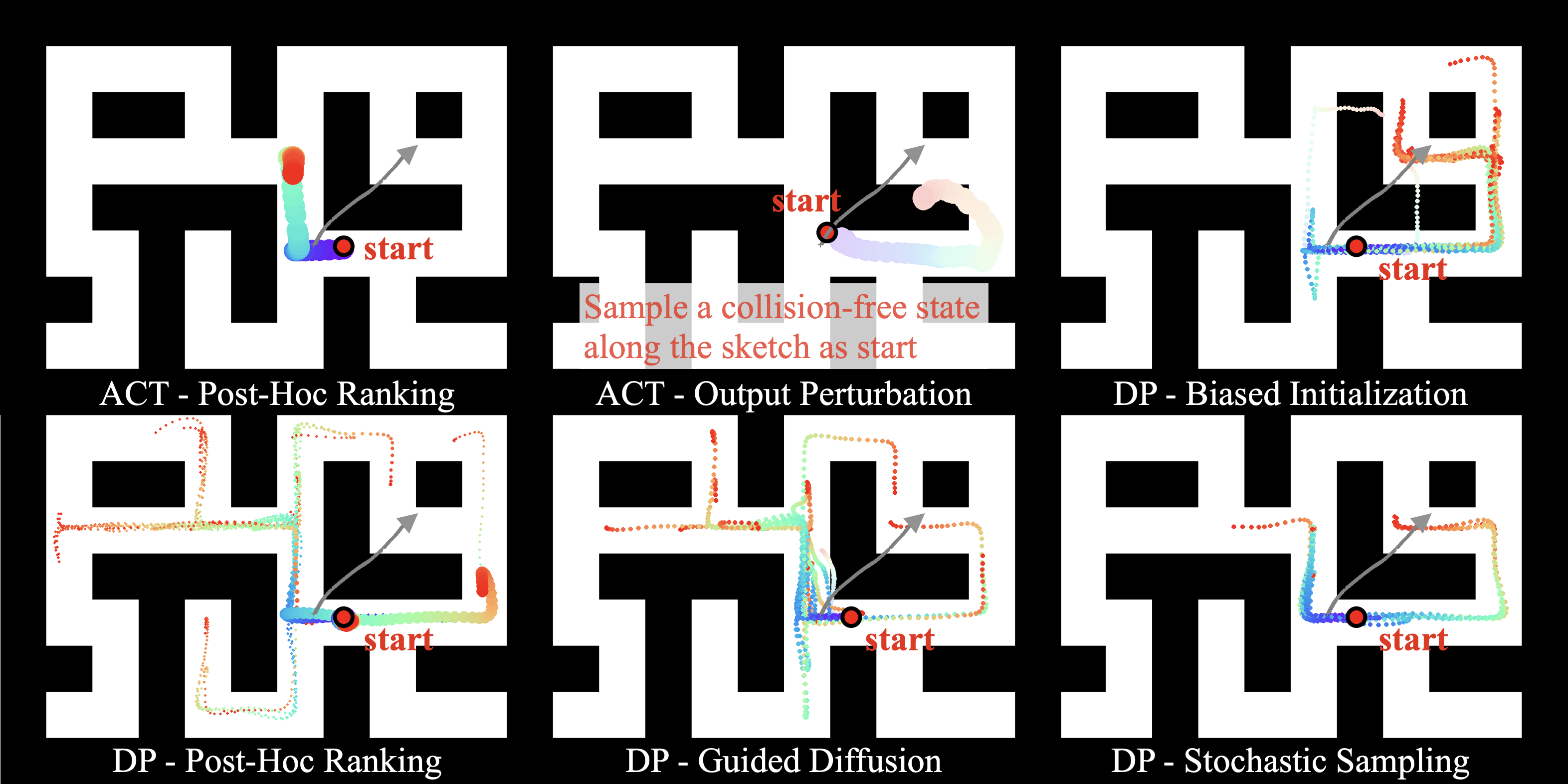}
    \caption{\small \textbf{Maze2D Qualitative Comparisons.} We visualize trajectories (color-coded from blue to red over time) sampled with various steering methods from two policy classes (ACT and DP) given a sketch in gray. Trajectory thickness reflects similarity to the sketch after ranking, and samples in collision are tinted white. \SGS{} preserves collision-free constraints while aligning with user intent.}
    \label{fig:maze2d_qualitative}
    
  \end{minipage} \hfill
  \begin{minipage}[t]{0.38\textwidth} 
  \vspace*{-152pt}
    \centering
    \includegraphics[width=0.90\linewidth]{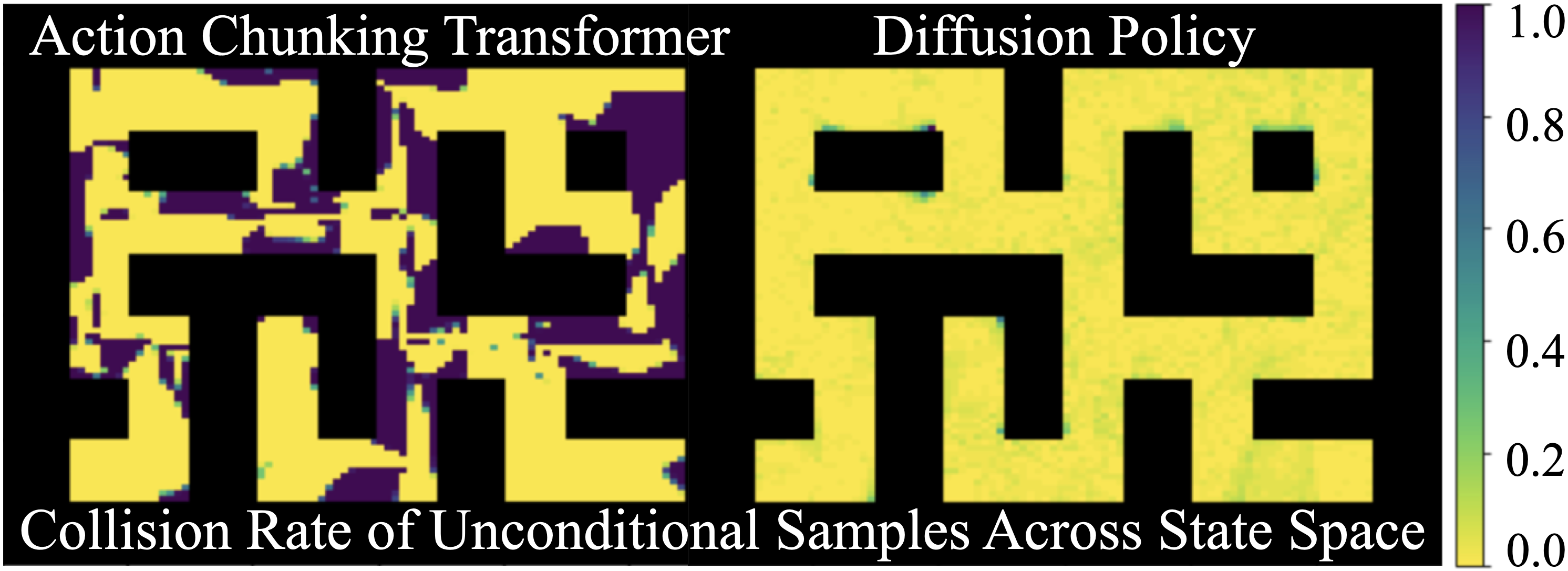}
    \captionsetup{width=0.9\linewidth}
    \caption{\small \textbf{Robustness of ACT/DP in Maze2D.}\\}
    \label{fig:maze_csr}
    \vspace{8pt}
    \captionsetup{width=0.99\linewidth}
        \scriptsize
        \centering
        \setlength{\tabcolsep}{3pt} 
        \begin{tabular}{llcccccc}
        \toprule
        & & \textbf{\RS} & \textbf{\PR} & \textbf{\OP} & \textbf{\BI} & \textbf{\GD} & \textbf{\SGS} \\
        \midrule
        \multirow{3}{*}{\textbf{ACT}} 
        & Min $\mathcal{L}_2$ $\downarrow$ & 0.26 & 0.26 & 0.26 & - & - & - \\
        & Avg $\mathcal{L}_2$ $\downarrow$ & 0.26 & 0.26 & 0.26 & - & - & - \\
        & Collision $\downarrow$ & 0.16 & 0.16 & 0.35 & - & - & - \\
        \midrule
        \multirow{3}{*}{\textbf{DP}} 
        & Min $\mathcal{L}_2$ $\downarrow$ & 0.27 & 0.16 & 0.16 & 0.11 & 0.11 & \textbf{0.10} \\
        & Avg $\mathcal{L}_2$ $\downarrow$ & 0.28 & 0.28 & 0.28 & 0.14 & 0.18 & \textbf{0.12} \\
        & Collision $\downarrow$ & 0.01 & 0.01 & 0.02 & 0.06 & 0.06 & \textbf{0.01} \\
        \bottomrule
        \end{tabular}
        \captionof{table}{\small \textbf{Maze2D Results.} Mean collision rate and $\mathcal{L}_2$ distance between $\mathbf{z^\text{sketch}}$ and the closest sample (min) / all samples (ave) per batch across trials. \SGS{} achieves the best alignment with minimal collisions.}
        \label{tab:maze}
  \end{minipage}
  \vspace{-15pt}
\end{figure*}

\textbf{Stochastic Sampling (\SGS).}
Finally, in \textit{Stochastic Sampling}, we use annealed MCMC to optimize the composition of the diffusion model $\pi_\theta$ and the objective $\xi(\tau_i, \mathbf{z})$~\cite{du2023reduce}. Here, the denoising function $\epsilon_\theta(\tau_i, i)$ at each timestep $i$ represents the score $\nabla_\tau \log p_i(\tau)$ for a sequence of probability distributions $\{p_i(\tau)\}_{0\le i\le N}$, where $p_N(\tau)$ is Gaussian and $p_0(\tau)$ is the distribution of valid trajectories in the environment. Simultaneously, the objective $\xi(\tau, \mathbf{z})$ defines an energy-based model (EBM) distribution $q(\tau) \propto e^{-\xi(\tau, \mathbf{z})}$. Steering toward user intent then corresponds to sequentially sampling from $p_N(\tau) q(\tau)$ to $p_0(\tau) q(\tau)$, yielding final samples from $p_0(\tau) q(\tau)$ that are both valid within the environment and minimize the specified objective.

We implement this sequential sampling using the annealed ULA MCMC sampler, which can be implemented in a similar form to the guided diffusion code~\cite{du2023reduce}. First, we initialize a noisy trajectory $\tau_N \sim \mathcal{N}(0, I)$, corresponding to a sample from $p_N(\tau) q(\tau)$. We then run $M$ steps of MCMC sampling at timestep $i$ using the update equation:
\begin{equation}
\tau_{i} = \tau_i - \gamma_i (\epsilon_\theta(\tau_i, i) + \beta_i \nabla_{\tau_i} \xi(\tau_i, \mathbf{z})) + \sigma_i \eta,
\label{eqn:mcmc_no_contract}
\end{equation}
repeated $M-1$ times, followed by a final reverse step in Equation \ref{eqn:guided_diffusion}
to obtain a sample $\tau_{i-1}$ from $p_{i-1}(\tau) q(\tau)$. These steps closely resemble reverse sampling in Equation~\ref{eqn:guided_diffusion} and can be implemented by modifying four lines in the guided diffusion code (Algorithm~\ref{alg:psuedocode}). To implement the sampling of Equation~\ref{eqn:mcmc_no_contract}, we take the intermediate clean trajectory prediction $\tilde{\tau}_0$ obtained via reverse sampling on $\tau_i$, followed by a forward diffusion step with noise level $i$ to update $\tau_{i}$. 
The addition of multiple reverse sampling steps at a fixed noise level better approximates sampling from a product distribution, as shown in Figure~\ref{fig:gd_vs_ss}, producing samples that satisfy likelihood constraints and user objectives. 
Across our experiments, \SGS{} provides the most proficient policy steering.



\section{Experiments}

We evaluate the effectiveness of inference-time steering methods in improving continuous \textbf{M}otion \textbf{A}lignment (\texttt{MA}) in Maze2D and discrete \textbf{T}ask \textbf{A}lignment (\texttt{TA}) in the Block Stacking and Real World Kitchen Rearrangement tasks. Additionally, we report how steering affect \textbf{C}onstraint \textbf{S}atisfaction (\texttt{CS}) among samples.

\subsection{Maze2D - Continuous Motion Alignment (\texttt{MA})} 

For continuous motion alignment, we use Maze2D \cite{fu2020d4rl} to evaluate whether a generative policy trained exclusively on collision-free navigation demonstrations can remain on a collision-free motion manifold when steered with sketches that violate constraints. To test the impact of the pre-trained policy class, we train a VAE-based action chunking transformer (ACT) \cite{zhao2023learning} and a diffusion policy (DP) \cite{chi2023diffusion} on 4 million navigation steps between random locations in a maze environment. DP is trained with a DDIM \cite{song2020denoising} scheduler over 100 training steps $(N=100)$. The training objective focuses solely on modeling the data distribution (i.e., collision-free random walk) without any goal-oriented objectives.

At inference time, a given policy is kept frozen to benchmark various steering methods. We generate 100 random locations in the maze, each paired with a user sketch $\mathbf{z}^{\text{sketch}}$ that may not be collision-free. These sketch inputs steer the generation of a batch of 32 trajectories per trial from the policy. For DP, the scheduler is allocated 10 inference steps, with a guide ratio of $\beta_{i \leq N} = 20$ for \GD\ and $\beta_{i \leq N} = 60$ for \SGS\, where the MCMC sampling steps are set to $M=4$. To incorporate $\mathbf{z}^{\text{sketch}}$ in the \OP\ sampling procedure, an early portion of the sketch is sampled to identify a non-collision state, resetting the starting location accordingly. To evaluate steering, we report the collision rate ($1 - \texttt{CS}$) and the $\mathcal{L}_2$ distance between the sketch and the closest trajectory (Min $\mathcal{L}_2$) or all trajectories (Avg $\mathcal{L}_2$) per batch, which measures negative \texttt{MA}. Min $\mathcal{L}_2$ shows the best alignment, while Avg $\mathcal{L}_2$ captures the overall distribution alignment after steering. 

\begin{figure*}[t] 
  \begin{minipage}[t]{0.60\textwidth} 
    \raggedleft
    \includegraphics[width=\linewidth]{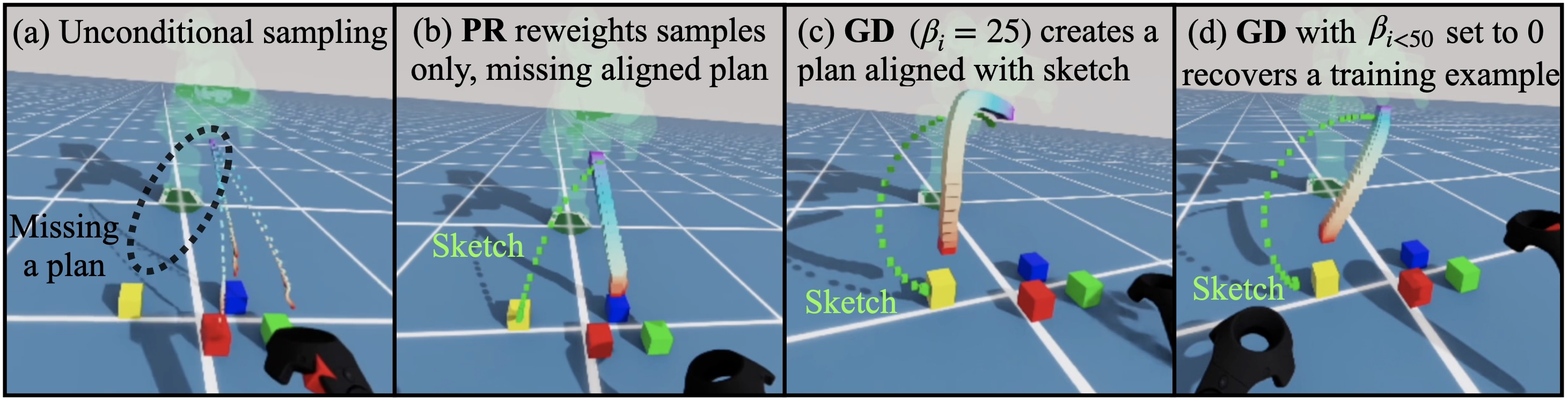} 
      \caption{\small \textbf{Block Stacking Qualitative Comparisons.} (a) Unconditional sampling from a DP may miss intended plans, which (b) \PR{} cannot recover, but (c) \GD{} can. (d) Adjusting the number of diffusion steps with steering (set guide ratio $\beta_i = 0$) balances similarity to the sketch versus adherence to the training distribution.}
      \label{fig:block}
  \end{minipage} \hfill
  \begin{minipage}[t]{0.37\textwidth} 
  \vspace*{-76pt}
    \small
    \captionsetup{width=\linewidth}
    \scriptsize
\setlength{\tabcolsep}{0pt} 
\begin{tabularx}{\linewidth}{@{}l*{3}{>{\centering\arraybackslash}X}@{}}
\toprule
\textbf{Method Type (DP)} & \textbf{\PR} & \textbf{\GD} ($\beta_{i<50}$=$0$) & \textbf{\GD} ($\beta_{i}$=$100$) \\
\midrule
\tikzmark{TA}TA (Alignment: AS+AF)       & 33\% & 83\% & 86\% \\
\tikzmark{CS}CS (Success: AS+MS)         & 100\% & 84\% & 15\% \\
\tikzmark{AS}\textbf{Aligned Success (AS)} & 33\% & \textbf{67\%} & 15\% \\
\tikzmark{AF}Aligned Failure (AF)          & 0\% & 16\% & 71\% \\
\tikzmark{MS}Misaligned Success (MS)       & 67\% & 17\% & 0\% \\
\tikzmark{MF}Misaligned Failure (MF)       & 0\% & 0\% & 14\% \\
\bottomrule
\end{tabularx}
\captionof{table}{\small \textbf{Block Stacking Results.} \texttt{TA} is the percentage of interactions that achieve aligned execution, regardless of outcome. \texttt{CS} is the percentage of picking/placing success, regardless of alignment.}
\label{tab:block}
  \end{minipage}
    \vspace*{-15pt}
\end{figure*}

Our findings, illustrated in Figure \ref{fig:maze_multi_trend}, reveal a tradeoff between alignment and constraint satisfaction. Specifically, aggressive steering improves \texttt{MA} but reduces \texttt{CS} and increases collisions. Additionally, a policy with multimodal predictions (DP) combined with \PR\ effectively improves alignment without exacerbating distribution shift. However, if the intended plan is absent from the initial sampled batch, \PR\ cannot discover it (Figure \ref{fig:maze2d_qualitative}). In contrast, a policy with unimodal predictions (ACT) cannot be steered to improve alignment with \PR. If the policy lacks robustness (Figure \ref{fig:maze_csr}), \OP\ can introduce significant distribution shift. Finally, diffusion-specific steering methods can transform constraint-violating sketches into the nearest collision-free samples on the data manifold. Among these, \SGS\ achieves the best \texttt{MA} and \texttt{CS} tradeoff, as shown in Table \ref{tab:maze} and Figure \ref{fig:maze2d_qualitative}.
 

\subsection{Block Stacking - Discrete Task Alignment (\texttt{TA})}

We evaluate discrete task alignment by testing whether a multistep generative policy, with multimodal predictions at each step, can be steered to solve a long-horizon task following a user-preferred execution sequence. For this, we design a 4-block stacking domain in the Isaac Sim environment \cite{mittal2023orbit}. The simulation initializes four blocks at random positions, and motion trajectories are generated using CuRobo \cite{sundaralingam2023curobo}. The planner randomly selects blocks to pick and place, sometimes disassembling partial towers to rebuild them elsewhere. We train a DP (DDIM with $N=100$) on 5 million steps from this dataset to learn a motion manifold of valid pick-and-place actions without goal-oriented behavior. As shown in Figure \ref{fig:block}(a), the learned policy exhibits multimodality across a discrete set of trajectories.

At inference time, we steer the policy to achieve a specific stacking sequence, completing a 4-block tower. To facilitate 3D steering, we develop a virtual reality (VR)-based system that allows users to provide 3D sketches within the simulation environment. 
In each interaction trial, the user observes the policy's unconditional rollouts before providing a sketch for conditional sampling. If the conditional sample with the smallest $\mathcal{L}_2$ distance to the sketch corresponds to the intended block, the trial is considered successfully aligned. If the policy execution also succeeds, the trial is deemed constraint-satisfying. We report \texttt{TA} and \texttt{CS} across interaction trials for \PR{} and \GD{} with $\beta_{i \leq N}=25$ in Table \ref{tab:block}. Again, we see that higher \texttt{TA} correlates with lower \texttt{CS}.

Additionally, we experiment with a strategy to mitigate distribution shift during sampling with \GD{}. Rather than keeping the guide ratio $\beta_i$ constant for all $i = N, \dots, 1$, we deactivate steering by setting $\beta_{i \leq I} = 0$ for later steps. This approach aligns the low-frequency component of the noisy sample with user input in early diffusion steps while reverting to unconditional sampling after step $I$. Figure \ref{fig:block}(c-d) demonstrate that the original \GD{} produces a curved trajectory resembling the sketch, while the modified \GD{} ($I = 50$) retrieves a straight-line trajectory from the CuRobo training dataset with the correct discrete alignment.

\subsection{Real World Kitchen - Discrete Task Alignment (\texttt{TA})}

\begin{figure*}[t!] 
  \begin{minipage}[t]{0.21\linewidth}
    \raggedright
    \includegraphics[width=0.97\linewidth]{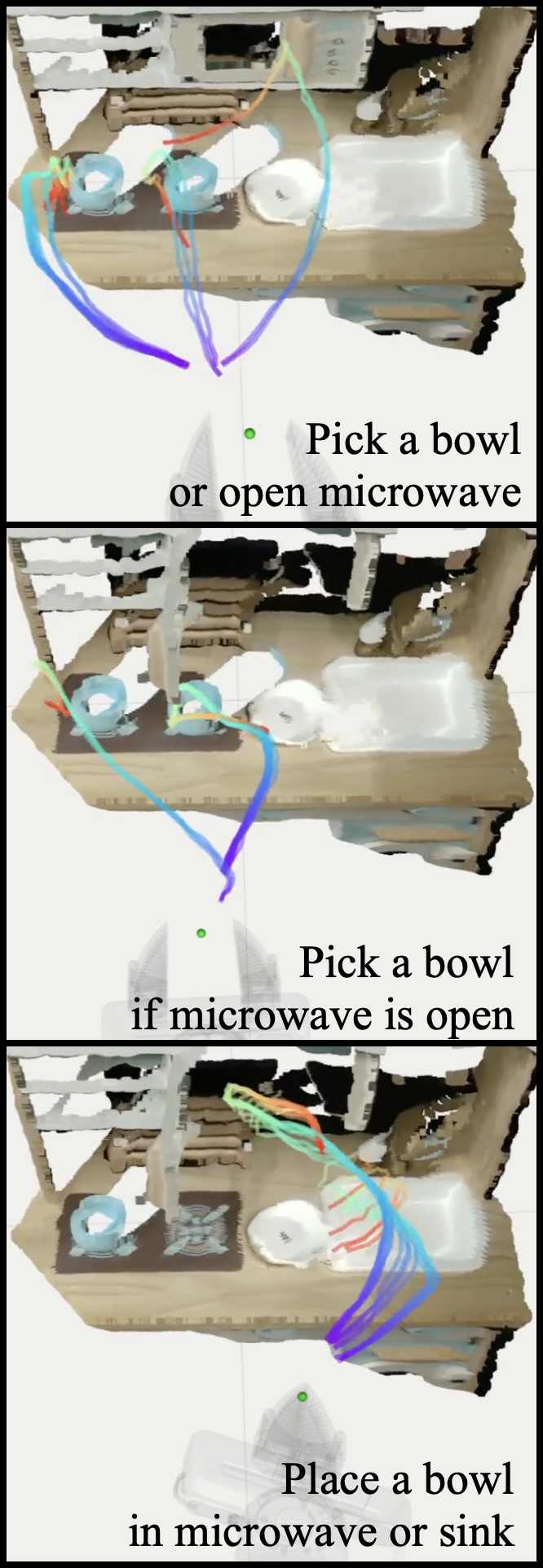} 
    \caption{\small \textbf{Multimodal Skills.}}
    \label{fig:kitchen_multimodal}
  \end{minipage}
  \hfill 
  \begin{minipage}[t]{0.79\linewidth}
    \raggedleft
    \vspace{-298pt}
    \begin{minipage}[t]{.98\linewidth}
        \includegraphics[width=\linewidth]{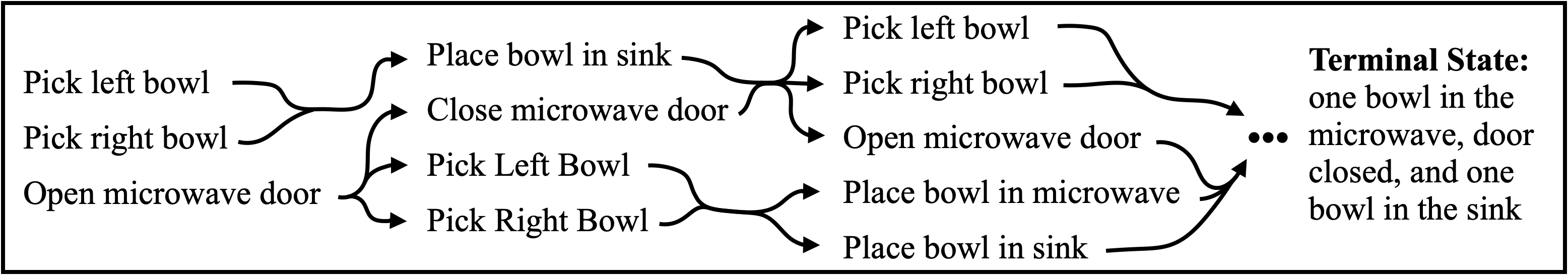}
        \caption{\small \textbf{Multimodal Valid Sequence for Kitchen Cleaning.} Steering selects a preferred legal sequence of skills to be executed until the terminal state is reached. This task requires a minimum of six steps.}
        \label{fig:kitchen_tree}
        \vspace{5pt} 
    \end{minipage}
    
    \begin{minipage}[t]{.98\linewidth}
        \includegraphics[width=\linewidth]{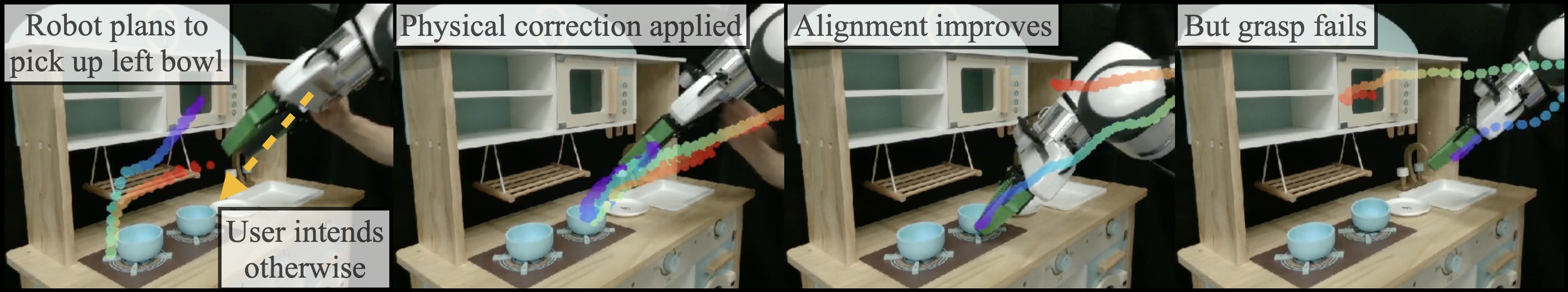}
        \caption{\small \textbf{Tradeoff Between Alignment and Distribution Shift.} As the user steers the policy to align with their intent, inference-time interactions may exacerbate distribution shift and lead to execution failure.}
        \label{fig:kitchen_rollout}
        \vspace{5pt} 
    \end{minipage}

    \begin{minipage}[t]{0.98\linewidth}
        \includegraphics[width=\linewidth]{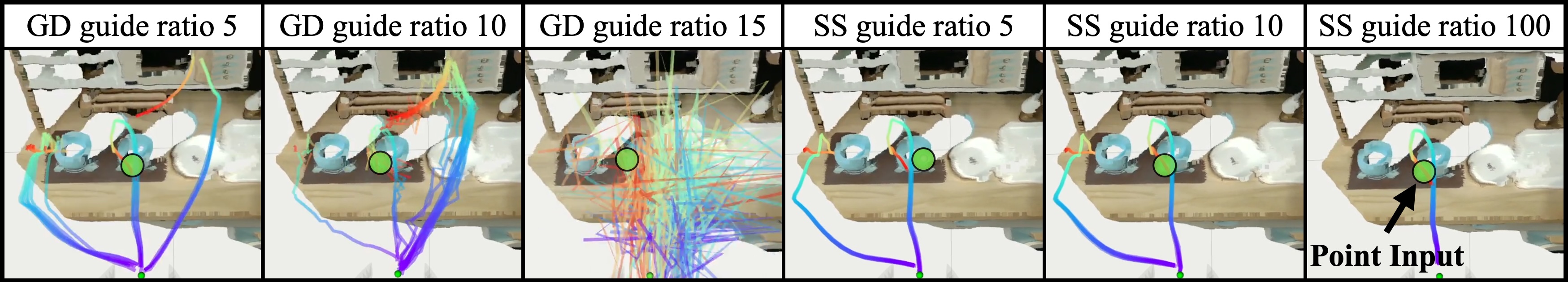}
        \caption{\small \textbf{Sensitivity to Guide Ratio $\beta_i$.} When $\beta_i$ is small, steering (via point input in this case) is ineffective for both \GD\ and \SGS. As $\xi_i$ increases, \GD\ begins to produce incoherent trajectories, while \SGS\ successfully identifies the intended skill. The same $\beta_i$ is applied for all $i \leq N$ ($N=100$).}
        \label{fig:kitchen_ablation}
    \end{minipage}
  \end{minipage}
    \vspace{-10pt}
\end{figure*}

To evaluate inference-time steering of multistep, multimodal policies in a real-world setting, we construct a toy kitchen environment and generate demonstrations using kinesthetic teaching. We focus on two tasks: (1) placing a bowl in the microwave and (2) placing a bowl in the sink. For each task, we collect 60 demonstrations and combine them into a dataset to train a diffusion policy (DP) over 40,000 steps. Figure \ref{fig:kitchen_multimodal} illustrates that the learned motion manifold exhibits distinct multimodal skills based on the end-effector pose and gripper state. Unlike the block stacking experiment, merging datasets from different tasks introduces scenarios where skill sequences are not feasible—for example, placing a bowl in the microwave before opening the microwave door. Therefore, in this context, the \texttt{CS} metric not only measures the success of individual skills but also evaluates whether the resulting sequence is valid as shown in Figure \ref{fig:kitchen_tree}.

At inference time, users can steer execution towards a preferred, valid sequence by clicking a pixel in the scene camera view to specify the intended skill. The corresponding 3D location of the pixel is visualized with a red sphere that turns green upon activation of the steering input. We also experiment with physical corrections to the end-effector pose to trigger behavior switches, but as shown in Figure \ref{fig:kitchen_rollout}, this type of interaction often leads to execution failures. 

We evaluate the effectiveness of \GD, \SGS, and \OP{} in enabling users to achieve specific sequences of discrete skills. During real-time policy rollouts (7 Hz), users observe a randomly sampled skill and select a different one through interactions. We report whether the interaction successfully causes the intended behavior switch and whether it results in successful execution in Table \ref{tab:kitchen}. For \GD, we use a guide ratio $\beta_i=5$ for all diffusion steps ($N=100$), while for \SGS{}, $\beta_i=100$ is used. These choices are based on the observation that increasing the guide ratio for \GD{ }disrupts the diffusion process without improving alignment (Figure \ref{fig:kitchen_ablation}). In contrast, higher guide ratios for \SGS{} enhance alignment without producing noisy trajectories. Thus, \GD{} with $\beta_i=5$ serves as a baseline for weak steering, while \OP{}—allowing users to physically correct the robot end-effector trajectory during execution—functions as an aggressive steering baseline. Both \GD{} and \SGS{} are steered with pixel inputs. In Table \ref{tab:kitchen}, as alignment \texttt{TA} increases, the constraint satisfaction rate \texttt{CS} decreases. The best steering method (\SGS) has a higher failure rate than rolling out randomly (\RS) but improves \textbf{Aligned Success by 21\%} without any fine-tuning.

\section{Related works}

\begin{table}[t]
\centering
\setlength{\tabcolsep}{0pt} 
\begin{tabularx}{0.8\linewidth}{@{}l*{4}{>{\centering\arraybackslash}X}@{}}
\toprule
\textbf{Method Type (DP)} & \textbf{\RS} & \textbf{\GD} & \textbf{\SGS} & \textbf{\OP} \\
\midrule
\textbf{Interaction Type} & \textbf{N/A} & \textbf{Point} & \textbf{Point} & \textbf{Correction} \\
\midrule
\tikzmark{ta}TA (Alignment: AS+AF) & 38\% & 37\% & 71\% & 89\% \\
\tikzmark{cs}CS (Success: AS+MS) & 90\% & 82\% & 73\% & 37\% \\
\tikzmark{as}\textbf{Aligned Success (AS)} & 34\% & 32\% & \textbf{55\%} & 30\% \\
\tikzmark{af}Aligned Failure (AF) & 4\% & 5\% & 16\% & 59\% \\
\tikzmark{ms}Misaligned Success (MS) & 56\% & 50\% & 18\% & 7\% \\
\tikzmark{mf}Misaligned Failure (MF) & 6\% & 13\% & 11\% & 4\% \\
\bottomrule
\end{tabularx}



\begin{tikzpicture}[overlay, remember picture]
    \draw[decorate, decoration={brace, mirror, amplitude=3pt}, thick]
        ($(as)+(-0.1cm, 0.2cm)$) -- ($(mf)+(-0.1cm, 0.0cm)$)
        node[midway, left=0.1cm]{\footnotesize 100\%};
\end{tikzpicture}

\caption{\small \textbf{Real World Kitchen Results.} We evaluate whether a user can steer a policy to switch from a randomly sampled skill to an intended skill and maintain successful execution. Overall, as alignment (\texttt{TA}) improves, the success rate (\texttt{CS}) decreases.}
\label{tab:kitchen}
\vspace{-15pt}
\end{table}

\textbf{Learning for Human-Robot Interaction.} Recently, learning from demonstrations \cite{chi2023diffusion,zhao2023learning} has achieved significant success in robotic manipulation. Despite this progress, real-time human input is often absent during inference-time policy rollouts. To address this gap, natural human-robot interfaces \cite{perzanowski2001building,berg2020review} have been employed when deploying robots in human environments. Various input forms, such as language, sketches, and goals \cite{team2024octo,brohan2023can,sundaresan2024rt,ding2019goal,mees2022matters,shi2024yell}, have also been studied to convey human intent to robots. Inspired by \cite{yoneda2023noise, ng2023diffusion}, our framework repurposes pre-trained generative policies for HRI settings, accommodating real-time human input. In this work, we focus on physical interactions, as they often provide grounding information that complements language prompts.

\textbf{Learning from Human Demonstrations.} Generative modeling \cite{urain2024deep,chi2023diffusion,lee2024behavior} has advanced imitation learning from multimodal, long-horizon demonstrations, enabling dexterous skill acquisition. Diffusion models \cite{chi2023diffusion}, are particularly effective at capturing the multimodal nature of human demonstrations, with their implicit function representation allowing flexible composition with external probability distributions \cite{janner2022planning, liu2022compositional, du2023reduce}. Previous research has explored using latent plans to support long-horizon tasks \cite{wang2022hierarchical,zhao2023learning,lynch2020learning}, but these focus on demonstrations with a single, high-quality behavior mode. In this work, we focus on generative modeling of multiple behavior modes \cite{wang2022temporal}, which is essential for enabling user interactions that require policies to adapt to inputs at inference time.

\textbf{Inference-Time Behavior Synthesis.} In robotics, inference-time composition has been explored as a method for achieving structured generalization \cite{du2019model,janner2022planning,gkanatsios2023energy, yang2023compositional, reuss2023goal,mishra2023generative,urain2023se}. Approaches like BESO \cite{reuss2023goal} leverage learned score functions combined with classifier-free guidance to enable goal-conditioned behavior generation. Similarly, SE3 Diffusion Fields \cite{urain2023se} use learned cost functions to generate gradients for joint motion and grasping planning, while V-GPS \cite{nakamoto2024steering} employs a learned value function to guide a generalist policy through re-ranking. PoCo \cite{wang2024poco} synthesizes behavior across diverse domains, modalities, constraints, and tasks through gradient-based policy composition, supporting out-of-distribution generalization. Building on PoCo, our work investigates how different types of real-time physical interaction can effectively steer policy at inference time.

\section{CONCLUSION}
In this work, we propose the Inference-Time Policy Steering (ITPS) framework, which integrates real-time human interactions to control policy behaviors during inference without requiring explicit policy training. We demonstrate how this approach enables humans to steer policies and benchmark several algorithms across both simulation and real-world experiments.
One limitation of our work is the reliance on an expensive sampling procedure to produce behaviors aligned with human intent. In future work, we aim to distill the steering process into an interaction-conditioned policy to achieve faster responses to human interactions and conduct a user study to further validate steerability.

\addtolength{\textheight}{-3cm}   
\clearpage





\section{Acknowledgment}

\noindent We would like to thank Michael Hagenow, Andreea Bobu, Phillip Isola, Jiayuan Mao, Leslie Kaelbling, Chuer Pan, Cheng Chi, and Sixian Wang for their invaluable advice and help.

\bibliographystyle{IEEEtran} 
\bibliography{IEEEabrv}

\end{document}